# Non-Standard Words as Features for Text Categorization

Slobodan Beliga, Sanda Martinčić-Ipšić

Department of Informatics
University of Rijeka
Radmile Matejčić 2, 51000 Rijeka, Croatia
{sbeliga, smarti}@inf.uniri.hr

**Abstract** - This paper presents categorization of Croatian texts using Non-Standard Words (NSW) as features. Non-Standard Words are: numbers, dates, acronyms, abbreviations, currency, etc. NSWs in Croatian language are determined according to Croatian NSW taxonomy. For the purpose of this research, 390 text documents were collected and formed the SKIPEZ collection with 6 classes: official, literary, informative, popular, educational and scientific. Text categorization experiment was conducted on three different representations of the SKIPEZ collection: in the first representation, the frequencies of NSWs are used as features; in the second representation, the statistic measures of NSWs (variance, coefficient of variation, standard deviation, etc.) are used as features; while the third representation combines the first two feature sets. Naive Bayes, CN2, C4.5, kNN, Classification Trees and Random Forest algorithms were used in text categorization experiments. The best categorization results are achieved using the first feature set (NSW frequencies) with the categorization accuracy of 87%. This suggests that the NSWs should be considered as features in highly inflectional languages, such as Croatian. NSW based features reduce the dimensionality of the feature space without standard lemmatization procedures, and therefore the bag-of-NSWs should be considered for further Croatian texts categorization experiments.

## I. Introduction

Categorization or classification is tasked to classify a given data instance into a pre-specified set of categories. In the domain of text categorization, the task is called text categorization (TC) – given a set of categories (subjects) and a collection of text documents, the process of finding the correct topics for each document [1]. Typically, texts are represented with vectors of word's frequencies in the bag-of-words model. The problem with this approach is the dimensionality of the features vectors, which is equal to the number of different words in the collection [1, 2].

One of the solutions for the dimensionality problem is **Feature Selection** which can be: heuristic, guided by linguistic and domain knowledge, or statistical [3].

In linguistically guided approach, the dimensionality of the features vector is reduced with stop words (*a*, *an*, *the*, *of*, etc.) omission and lemmatization. Lemmatization changes each inflected word form with its lemma (*walked*, *walks*, *walking* – with lemma *walk*) and therefore reduces the dimensionality of feature vectors, which is of essence in morphologically rich languages such as Croatian. Some feature selection approaches ignore terms that are too frequent or too rare according to empirically chosen thresholds [3]. As data sets become more complex, heuristics are usually not sufficient. Therefore, another solution is to use measures such as Information Gain and Chi-Square, to define relevance of each feature [4].

Another approach to dimensionality problem is **Dimensionality Reduction by Feature Extraction**. It implies a much smaller set of synthetic features from the original feature set, using synthetic features rather than naturally occurring words. This will avoid the problems inherited from natural language (polysemy, homonymy, etc.). Using word groups, instead of individual words as features, with a high degree of semantic relatedness, proved to be justified in clustering [1]. Systematic approach - Latent Semantic Indexing (LSI) is based on this principle [1].

In this paper we propose opposed approach based on Non-Standard Words (NSWs) as features. Instead of bag-of-words, we construct the bag-of-NSWs, and use them to train different text classifiers. The peculiarity of proposed approach is in using NSWs which would usually be omitted from the text during the text cleaning procedures. Non-standard words are: numbers, dates, time, abbreviations, acronyms, currency, measurement units, etc. Text normalization transforms NSWs in their extended form: the abbreviation */dr./* into */doctor/* or acronym */SMS/* into */Short Message Service/*.

The motivation for this work rises from an idea that discarded characters and numbers are sufficient for discrimination of classification of the text categories, reducing the feature vector dimensionality at the same time [2].

The paper is organized as follows: Section II. presents methodology of text normalization and categorization, Section III. presents the data collection, NSW features extraction procedures and training of classification models. The results are presented in the end of Section III. The paper concludes with discussion and some future work considerations.

This research is partially supported by the grant 13.13.2.2.07 of University of Rijeka.

## II. METODOLOGY AND RELATED WORK

### A. NSW extraction

Traditionally, in the field of natural language processing (NLP), normalization is the process of transforming the unit into its normal form (i.e. lemma). Normalization is also the first step in the text pre-processing of Text-to-Speech (TTS) systems. It is performed in the normalization module, which is responsible to identify a NSW token and to transform it into its expanded form [5, 6]. Text Normalization for Croatian Speech Synthesis is described in [7]. In this work, the Croatian NSW taxonomy driven rule based approach was used for NSWs extraction [7, 8].

Module for normalization initially detects the NSW and separates it from standard words. Then, detected NSW is transformed into its expanded form, suitable for the TTS system. In this experiment, the process of transforming NSW to expanded form is not required. Instead, we detect each NSW and determine its NSW type as is suggested in taxonomy for classification of non-standard words in Croatian language [8]. In addition, every occurrence of NSW for each NSW type is counted. Our approach identifies NSWs and then normalizes detected NSWs following the taxonomy of Croatian NSWs by combining programmed rules and the lookup dictionary.

### B. Text categorization

Text categorization is tasked with assigning one of predefined classes to a document. Usually text categorization is performed using a bag-of-words model borrowed from information retrieval (IR), and most often based on TF-IDF model [1-3]. Documents are represented as feature vectors, which are, according to the bag-of-words model, unordered list of words with the frequency of their occurrence in a document. Inverse document frequency is used to model the rareness of terms in a document collection. Feature vectors are used to train the classifier, usually Bayesian classifier, k-nearest neighbor (kNN), classification tree or support vector machines (SVM), etc. [1, 3, 9]. Some of the techniques for the vector dimensionality or the feature space reduction are: *stop words* and NSW removal [10], lemmatization or morphological normalization [11] - NLP based principles; selecting the most discriminative subset of features (document frequency, information gain, mutual information, odds ratio, etc.) – called feature selection [4, 12] - or deriving new smaller set calculated from original features – called feature extraction [13].

Our approach is presented in Figure 1.: in the first step - we detect all NSWs from the text; in the second step – for detected NSWs we assign its type (number, abbreviation, date, etc.); in the third step – we construct the bag-of-NSWs; in the fourth step – we train the classifiers. Finally, as a result of the text categorization we obtain the texts which are organized as categorized collection of texts – every text is associated with one of the possible categories.

## III. NSW BASED TEXT CATEGORIZATION

### A. Data (SKIPEZ)

For the purposes of normalization and categorization tasks we prepared the collection of texts in Croatian language – SKIPEZ (Službeno, Književno, Informativno, Popularno, Edukativno, Znanstveno). SKIPEZ contains 390 texts organized in 6 predefined classes: official, literature, informative, popular, educational and scientific. The distribution of text over classes is balanced so each class contains exactly 65 texts. Texts were selected according to their relevance to the class and according to the % of contained NSWs, because we needed a balance between both aspects in order to train a classifier. SKIPEZ statistics is shown in Table I., and additional details about SKIPEZ are in [14].

The content of texts in each subclass was carefully selected, in order to capture as many different and representative text in each class used and contain representative set of NSWs, in order to compile the exhausted list of Croatian NSWs is shown in Table II.

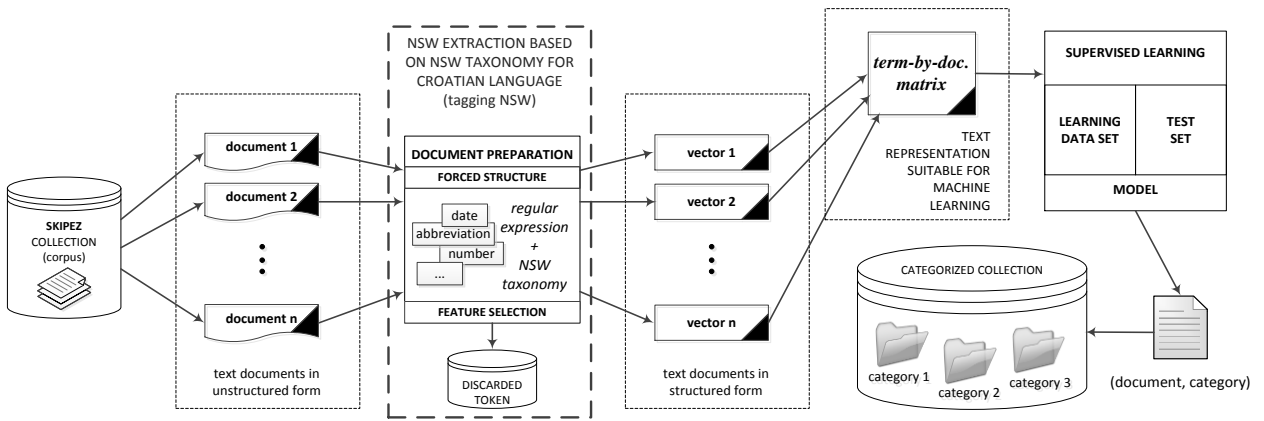

Figure 1. Shematic representation of NSW based text categorization

TABLE I.  NUMBER OF TOKENS* AND NON-STANDARD WORDS IN SKIPEZ CORPUS SHOWN PER 6 CATEGORY

| CATEGORY | Statistics of SKIPEZ corpus | | |
|---|---|---|---|
| | #TOKENS | #NSWs | %NSWs |
| official | 328.322 | 81.667 | 24,87 % |
| literature | 789.555 | 18.150 | 2,30 % |
| informative | 86.819 | 12.641 | 14,56 % |
| popular | 224.518 | 18.033 | 8,03 % |
| educational | 198.661 | 11.598 | 5,84 % |
| scientific | 644.571 | 73.979 | 11,48 % |
| **OVERALL** | **2.272.446** | **216.068** | **9,51 %** |

*(e.g. words and punctuation marks)

TABLE II.  NUMBER OF TEXT ACCORDING TO THE TEXT TYPE IN EACH SKIPEZ CATEGORY

| CATEGORY | text type/origin | #texts | |
|---|---|---|---|
| official | stock reports | 16 | 65 |
| | announc. ads | 10 | |
| | legislation gazette | 12 | |
| | statistics | 8 | |
| | public tenders | 9 | |
| | exchange rates | 10 | |
| literature | lyrics | 2 | 65 |
| | novel, story play | 17 | |
| | publicists | 31 | |
| | history, theology | 15 | |
| informative | traffic, traveling | 19 | 65 |
| | phone books | 6 | |
| | TV guides | 12 | |
| | weather forecasts | 18 | |
| | commercial adds | 10 | |
| popular | forums | 8 | 65 |
| | horoscope | 10 | |
| | recipes | 14 | |
| | lifestyle, cars, health | 12 | |
| | on-line sales catalogs | 10 | |
| | blogs | 6 | |
| | social network comm. | 5 | |
| educational | school books nature | 16 | 65 |
| | school books social | 15 | |
| | school books human. | 17 | |
| | school books technol. | 17 | |
| scientific | encyclopedia | 10 | 65 |
| | atlas | 9 | |
| | dictionaries | 9 | |
| | master and phd thesis | 15 | |
| | books | 10 | |
| | journal papers | 12 | |
| **OVERALL** | | | **390** |

*B. NSW Extraction Procedure*

The NSWs were extracted from the SKIPEZ text collection according to the approach presented in [7, 8]. Here, we only detected NSWs types according to this taxonomy, resulting in feature vectors consisting of NSWs types frequencies for each document in the collection. A feature vector for each text document in collection contains 85 values (calculated from NSW types frequencies):

- STRING part - 15 values: ordinal and nominal roman numbers, simple or compound abbreviations, abbreviations without full stop, chemical elements, measurement units, currency an monetary units, acronyms, inflected and special acronyms, symbols, emoticons and suffixes.

- NUMBER part - 21 values: numeric format date, time period, time, proportion, dimensions, short references, short and long telephone number, interval, positive and negative decimal numbers, decimal interval, positive and negative fractions, exponents, ordinal and nominal numbers, ordinal and nominal intervals, positive and negative nominal numbers, ISDN/UDK numbers.

- COMBINED part - 20 values: 3 different date formats, geographical coordinates, long references, biblical references, enumerations of tables, figures and titles, road designation, TV show label, traffic abbreviations, compound measurement units, vitamins, house numbers, chemical and mathematical formulas, program code, registration number, argo-emoticons, domain specific NSWs, and NSWs of unknown type.

- DERIVED part - 29 values: total of NSWs in the number class, total of NSWs in the string class, total of NSWs in combined class, total of NSWs in text, total of words in text, number of different NSW types, number of empty NSW types, ratio between filled features and subclasses, ratio between filled and empty NSW types, coefficient as number of NSW divided with number of words in text. Moreover, there are also features that summarize the total number of instances of certain NSWs subclasses that are inter-related and can be classified in the same superclass. These are respectively: total of full date, total of incomplete date, total of references, total of numeration, total telephone numbers, total of range, total of decimal numbers, total number of fractions, total number of road designations, total number of electronic addresses, total of ordinal numbers, total of nominal numbers, total of roman numbers, total number of abbreviations, total number of measurement units, total number of acronyms, total number of office designation, total number of formulas and total number of jargon terms.

The first representation of SKIPEZ collection is a feature set composed from NSW type frequencies. For each document the features vector contains 85 values (as listed above).

The second representation of text, based on the NSW frequencies, represents the base from which the values of the features are derived. In this case, the statistics serves as a tool for obtaining metadata from the previously calculated frequency of NSW occurrence. For each element of the population, therefore each vector, we calculate some of its numerical characteristics called statistical features [15, 16]. The main idea when choosing a feature vector is utilizing the basic statistical features, which can describe the dispersion of identified patterns of NSW in a particular text.

The statistical characteristics of each population (each vector) were counted. We observed the vector elements as a sequence of elements and determined the individual values of absolute and relative measures of dispersion: the arithmetic mean, the range of variation, standard deviation, variance, coefficient of variation, and also a measure of flatness and a measure of asymmetry. We then calculated the statistical characteristics exclusively for superclasses (string, number, and string + number) – higher classes from taxonomy (Figure 2.). This second group of calculations observes dispersion patterns within the superclass, i.e. numerically describes the statistical characteristics of the NSW dispersion in three superclasses. We calculated the following characteristics: the arithmetic mean of superclass, interquartile q3, interquartile q1, interquartile Iq, coefficient of quartile deviation and coefficient of variation.

The third representation of texts is formed as a union from the features of the first and second representations, therefore includes all their values. Through the combination of features, we try to take advantage of the knowledge that was previously extracted from all the texts, on the principle "the more, the better".

In experiments with all 3 feature sets we conduct an iterative process of adding the created vectors for each text, we gradually form a bag-of-NSWs model (*term-by-document matrix*) - see Figure 1. Since we used supervised machine learning, during the bag-of-NSWs construction to each feature vector we assign its actual category (see Table II.).

### C. Categorization Results

We trained the Naive Bayes, Classification Tree, kNN, CN2, C4.5, and Random Forest using the Orange [17] toolkit. As the measure of classification performance we used accuracy (1) [9].

$$\text{Accuracy} = \frac{Number\ of\ correct\ classifications}{Total\ number\ of\ test\ cases} \quad (1)$$

where a correct classification means that the learned model predicts the same class as the original class of the test case.

The results of trained classifiers (Naïve Bayes, Classification Tree, kNN, CN2, C4.5 and Random Fortest) are obtained for three different feature representations using 5 fold cross-validation. The text categorization results using the first feature set - NSW frequencies are shown in Table III.; using statistical features in Table IV. and for combination of features in Table V.

TABLE III. RESULTS OF TEXT CATEGORISATION WHERE TEXTS ARE REPRESENTED BY FREQUENCIES OF NSWs

| CLASSIFIER | #tokens | Accuracy [%] |
|---|---|---|
| Naïve Bayes | 263 | 67,44 |
| Classification Tree | 305 | 78,21 |
| kNN | 296 | 75,90 |
| CN2 | 306 | 78,46 |
| C4.5 | 309 | 79,23 |
| **Random Forest** | **335** | **86,15** |
| The mean score of all classifiers: | | 77,56 |

TABLE IV. RESULTS OF TEXT CATEGORISATION WHERE TEXTS ARE REPRESENTED STATISTICAL FEATURES

| CLASSIFIER | #tokens | Accuracy [%] |
|---|---|---|
| Naïve Bayes | 176 | 45,13 |
| Classification Tree | 223 | 57,18 |
| kNN | 220 | 56,41 |
| CN2 | 212 | 54,36 |
| C4.5 | 229 | 58,72 |
| **Random Forest** | **260** | **67,95** |
| The mean score of all classifiers: | | 56,62 |

TABLE V. RESULTS OF TEXT CATEGORISATION WHERE TEXTS ARE REPRESENTED BY UNION OF NSWs FREQUENCIES AND STATISTICAL FEATURES

| CLASSIFIER | #tokens | Accuracy [%] |
|---|---|---|
| Naïve Bayes | 176 | 62,05 |
| Classification Tree | 223 | 76,92 |
| kNN | 220 | 70,00 |
| CN2 | 212 | 74,62 |
| C4.5 | 229 | 78,97 |
| **Random Forest** | **260** | **86,92** |
| The mean score of all classifiers: | | 74,91 |

The experimental results can be analyzed twofold: by accuracy of different classifiers and by accuracy with respect to different types of representations in collection - Figure 3. Random Forest achieves the highest accuracy and Naïve Bayes the lowest in all representation of the collection. The accuracy of all classifiers with statistical features is generally worse than the frequency features (Table IV.). With statistical features, Random Forest achieved a slightly higher accuracy (increase less than 1%). Chakrabarti, Yang and Pedersen in [4] state that numerous techniques of selection of features improve the performance of the classifier only marginally. This was confirmed also with our experiment, although the selection of features used non-standard features of the vector.

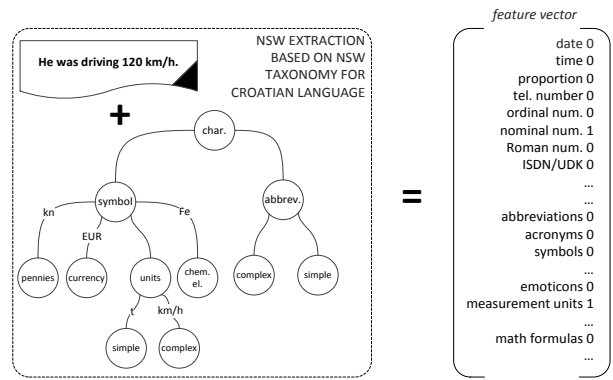

Figure 2. Forming a feature vector in Bag-of-NSWs

Experimental results can be analyzed according to text categories as well. Comparison of text categorization accuracy, only using Random Forest, is presented in Table VI. The highest accuracy was achieved for the class of informative texts, while the worst was achieved in the class of scientific and educational texts. The obtained results of classification leave room for discussion about the possibility of unification or reduction of scientific and educational domains because in this 2 categories is the highest number of misclassified texts. The science and education are closely related and they share the division of the subareas, fields and branches.

The worst scores for educational and scientific domain can be explained as an error that is probably caused by the fact that the educational and scientific texts are equally distributed in the scientific and artistic areas, fields and branches as required by the National Council for Science, and for Republic of Croatia Ministry of Science, Education and Sports, available at [18]. On the basis of this criterion texts are selected for SKIPEZ collection. Classifier failed to recognize the scientific style of writing. Encyclopedic texts, journals, articles and dissertations are apparently very similar to school textbooks, teaching materials and manuals through the NSW perspective.

## IV. DISCUSSION AND CONCLUSION

This paper presents categorization of Croatian texts using Non-Standard Words (NSW) as features. Results suggest that non-standard word forms can be used as features for representation of texts in text categorization. The associated NSW taxonomy is suitable form of background knowledge on the basis of which NSW can be automatically extracted from texts. It has been shown that the NSWs carry enough information about the nature of text, which is suitable for further classification. With this approach, it is possible to significantly reduce the dimensionality of feature vectors (space) and at the same time achieve good text categorization results. Using NSWs feature, the feature vectors have several times lower dimensions then the original ones and this way reduces the problem of sparse data.

Bag-of-NSWs approach is suitable for inflectional languages because extremely reduced dimensionality (only 85 features) which can avoid the procedure of lemmatization of words or morphological normalization.

Obtain results should be compared with standard TF-IDF classification method. Since the literature presents the support vector machine as successful, it would be appropriate to examine whether a classification using SVM provides better results than the Random Forest algorithm. In further work we plan to extend our work to the field of micro texts classification, since SMS and twitter texts contain a lot of NSWs.

TABLE VI. COMPARISON OF CATEGORIZATION ACCURACY WITH RESPECT TO 3 DIFFERENT FEATURE REPRESENTATIONS FOR RANDOM FOREST ALGORITHM

| CATEGORY | ACCURACY [%] | | | AVG [%]: |
|---|---|---|---|---|
| | *NSWs freq.* | *Statistics* | *NSWs freq. & Stat.* | |
| official | 76,92 | 66,15 | 81,54 | 74,87 |
| literature | 98,46 | 67,69 | 93,85 | 86,67 |
| informative | 100 | 83,08 | 100 | **94,36** |
| Popular | 100 | 66,15 | 92,31 | 86,15 |
| educational | 72,31 | 64,62 | 76,92 | **71,28** |
| scientific | 69,23 | 60,00 | 76,92 | **68,72** |
| AVG [%]: | 86,15 | 67,95 | **86,92** | |